\documentclass[conference]{IEEEtran}
\IEEEoverridecommandlockouts
\usepackage{cite}
\usepackage{amsmath,amssymb,amsfonts}
\usepackage{algorithmic}
\usepackage{graphicx}
\usepackage{textcomp}
\usepackage[table]{xcolor}

\usepackage{hyperref}

\usepackage{booktabs}
\usepackage{multirow}

\usepackage{xspace}

\usepackage{enumitem}
\usepackage{subcaption}

\usepackage{float}

\usepackage{flushend}

\usepackage[ruled,vlined]{algorithm2e}

\usepackage[skins,listings]{tcolorbox}
\def\BibTeX{{\rm B\kern-.05em{\sc i\kern-.025em b}\kern-.08em
    T\kern-.1667em\lower.7ex\hbox{E}\kern-.125emX}}
    
\usepackage{enumitem}
\usepackage{ifthen}
\usepackage{soul}   
\usepackage{cite}
    
\usepackage[frozencache=true,cachedir=.]{minted}

\usepackage{listings}

\newcommand{\etal}{\textit{et al.}\xspace}
\newcommand{\numpy}{\textsc{numpy}\xspace} %

\newcommand\exampleReferences[1]{\hyperref[lst:ex_stochastic]{{\color{green!50!black}{(\textsc{#1})}}}} %
\DeclareMathOperator*{\argmin}{argmin} 
\begin{document}
\pagenumbering{arabic} 
\pagestyle{plain}

\title{A Review and Refinement of Surprise Adequacy
\thanks{
This work was partially supported by the H2020 project PRECRIME,
funded under the ERC Advanced Grant 2017 Program (ERC Grant Agreement n. 787703).\newline
*Asterisks indicate both authors contributed equally\newline
Accepted at DeepTest 2021. 
© 2021 IEEE. Personal use of this material is permitted. Permission from IEEE must be
obtained for all other uses, in any current or future media, including
reprinting/republishing this material for advertising or promotional purposes, creating new
collective works, for resale or redistribution to servers or lists, or reuse of any copyrighted
component of this work in other works.
}
}

\author{
\IEEEauthorblockN{Michael Weiss*}
\IEEEauthorblockA{
\textit{Universit\`a della Svizzera italiana}\\
Lugano, Switzerland \\
michael.weiss@usi.ch}
\and
\IEEEauthorblockN{Rwiddhi Chakraborty*}
\IEEEauthorblockA{
\textit{Universit\`a della Svizzera italiana}\\
Lugano, Switzerland \\
rwiddhi.chakraborty@usi.ch}
\and
\IEEEauthorblockN{Paolo Tonella}
\IEEEauthorblockA{
\textit{Universit\`a della Svizzera italiana}\\
Lugano, Switzerland \\
paolo.tonella@usi.ch}
}

\maketitle

\begin{abstract}
    Surprise Adequacy (SA) is one of the emerging and most promising adequacy criteria for Deep Learning (DL) testing. 
    As an adequacy criterion, it has been used to assess the strength of DL test suites. In addition, it has also been used to find inputs to a Deep Neural Network (DNN) which were not sufficiently represented in the training data, or to select samples for DNN retraining. However, computation of the SA metric for a test suite can be prohibitively expensive, as it involves a quadratic number of distance calculations.
    Hence, we  developed and released a performance-optimized, but functionally equivalent, implementation of SA, reducing the evaluation time by up to 97\%. We also  propose refined variants of the SA computation algorithm, aiming to further increase the evaluation speed.
    We then performed an empirical study on MNIST, focused on the out-of-distribution detection capabilities of SA, which allowed us to reproduce parts of the results presented when SA was first released. The experiments show that our refined variants are substantially faster than plain SA, while producing comparable outcomes.
    Our experimental results exposed also an overlooked issue of SA: it can be highly sensitive to the non-determinism associated with the DNN training procedure.
\end{abstract}


\section{Introduction}
\label{sec:introduction}

Surprise Adequacy (SA) is a metric that has been proposed by Kim \etal \cite{Kim2018} in 2018
as part of \emph{Surprise Coverage}, a test adequacy criterion to assess the strength of a test set used for testing a deep neural network (DNN).
SA measures the \emph{surprise} caused by each test input to the DNN by comparing the inner state of the DNN to states observed in the DNN training data.

Besides being useful to measure the level of surprise coverage of a test set, SA has been shown to perform well on several other tasks,
among which \emph{out-of-distribution detection} (OODD) \cite{Kim2018} 
i.e., the detection of inputs to a DNNs which were not well represented in the training data used to train the DNN.
In fact, a highly surprising input is expected to be out-of-distribution with respect to the training data and the predictor's nominal data.
Finding OOD inputs is a crucial task in DL testing, as
inputs insufficiently represented during training are likely to lead to silent, undiscovered faults of the DNN (e.g., mis-classifications or vastly wrong predictions), which may cause failure of the overall DL system if no efficient countermeasures are implemented.
Thus, detecting OOD samples at runtime, as well as exercising them at testing time, is crucial to ensure the robustness of  DL based systems.
While there are other techniques to perform OODD in a DL based system,
such as autoencoders \cite{Stocco2020} or augmented DNN outputs \cite{Hendrycks2018},
SA aims at a different goal: quantifying and improving the adequacy of the test set.

As inputs which are surprising to a DNN are intuitively more useful for continued training of the DNN than unsurprising inputs, SA has proved to be a useful metric for selecting the data most useful in continuous training of DNNs\cite{Kim2018}. 

Alongside the original description of SA, its implementation was released under a permissive open source license\cite{Kim2018Code}.
Besides the wide applicability of SA, the availability of its implementation contributed to a large success of SA: Within two years since its publication, it has already been cited 105 times. 
Motivated by the popularity of SA in recent research, we provide a review and refinement of Surprise Adequacy in this paper. 
Specifically, we provide the following contributions:
\begin{description}[noitemsep]
\item [Usage Review] We reviewed the 105 citations of Kim \etal \cite{Kim2018}, the paper which first introduced SA, to find how and for which purpose SA is used in the literature.
\item [Improved Implementation] We conducted a review of the original implementation of SA and refactored it to improve both the SA runtime and its usability.
\item [Refined Approaches] We further tackled the high computational cost of SA by proposing several sampling strategies to be applied on the training dataset.
\item [Empirical Evaluation] We measured the performance of our improved implementation of the original approach and our refined, extended approaches on the MNIST dataset. 
\end{description}

\section{Surprise Adequacy}
\label{sec:surprise-adequacy}

SA is based on measuring the novelty of the DNNs inner state, i.e., the activations of its nodes, for a given input with respect to the inner states  observed for the training data.
In this paper, we consider the typical case where only the activations of a single layer are taken into account:
We denote the activations for any input $x$ in that layer as \emph{activation trace (AT)}, or formally $at(x)$, omitting the (constant) choice of the layer for simplicity.
Given a training set $T$ we can, for any input $x$, evaluate the novelty of $at(x)$ with respect to the training set ATs $\{at(x_t) \mid x_t \in T \}$. 
Specifically, Kim \etal \cite{Kim2018} propose two approaches to do so, respectively based on the \emph{likelihood} w.r.t. and the \emph{distance} from the training ATs respectively:

\subsection{Likelihood-Based SA (LSA)} LSA uses gaussian kernel density estimation (KDE) to learn the distribution of ATs. 
Given this estimated distribution, we can infer the probability of the AT of any novel input with respect to the learned distribution. Surprise is quantified as the negative log-likelihood of the probability density:
\[
LSA(x, T) = -\log(P(at(x))) = -\log(KDE(at(x)))
\]
    
\subsection{Distance-Based SA (DSA)} 
DSA, which is only applicable to classification problems, quantifies the level of surprise based on the distance of the AT of a given input to the training set ATs.
Specifically, surprise is quantified by measuring the ratio between the distance to the closest training AT of the same class, 
and the distance between the latter and  the closest training AT of any other class:
\[
DSA(x, T, M) = \frac{\lVert at(x) - at(t_1) \lVert}{ \min\limits_{t_2 \in T, M(t_2) \neq M(t_1)}\lVert at(t_1) - at(t_2) \lVert}
\]
where $M(i)$ denotes the class predicted by model $M$ for input $i$ and where $t_1 = \argmin\limits_{t \in T, M(t) = M(x)}\lVert at(x) - at(t) \lVert$ .

\subsection{Known Weaknesses}
Despite its popularity, SA and its two instances, LSA and DSA, suffer some  well known weaknesses.
First, using SA requires the choice of the layer(s) for which the ATs should be considered.
This choice can have a very high impact on the metric and its usages, such as  OODD.
Examples show that, even if the optimal choice of the layer would allow to have an OODD AUC-ROC of 0.99, the choice of a different layer in the same model can reduce performance to 0.82\cite{Kim2018}.
Second, calculating surprise for a set of test inputs does not scale well with the number of training samples:
For DSA, the distances of each test input to all ATs from the training set have to be computed, thus leading to a runtime proportional to the training set size and overall quadratic with the test set and training set size.
KDE, and thus LSA, is known not to scale well for large datasets either, with improvements and workarounds still being an open research question\cite{Raykar2010}.
Provided the very large training sets typically seen in problems where DNNs are the AI technique of choice,
SA tends to quickly become impractical if not applicable at all.

\section{Literature Review}

Within the $\sim$2 years since it's publication, SA had a large impact on research: 
As of January 15, 2021, Google Scholar lists 105 citations of the original SA  paper by Kim \etal \cite{Kim2018}.
In this section, we present a review of the usage of SA in these SA-citing papers.

Out of 105 citations, 18 were either invalid or duplicates and 4 were written in a non-English language. 
We removed them from our study, remaining with 83 valid unique papers citing SA.
Out of these papers, we found that 66 papers cite SA, typically in the related work or background sections, but do not use it in any experimental study.
The remaining 15 papers are most interesting to us, since
they use SA in some experiment.
In the remainder of this section, we will briefly mention the use of SA in each of these papers.\footnote{
A detailed list showing the classification of all papers in the categories mentioned in this paper is available as part of our replication package.
}

\subsection{Papers using SA as baseline metric}
SA has been used as a baseline metric to assess a variety of novel techniques:
\begin{itemize}
    \item DeepGini by Feng \etal \cite{Feng2020}, a method to evaluate prioritisation of tests to improve robustness of neural networks.
    \item DeepImportance by Gerasimou \etal \cite{Gerasimou2020}, an approach that evaluates deep learning system tests based on a new metric they introduce called ``importance''.
    \item MCP by Shen \etal \cite{Shen2020}, a method to improve neural network retraining by selecting a subset of test inputs.
    \item RAID by Eniser \etal \cite{Eniser2020}, a randomised adversarial input detection scheme.
    \item Robustness prediction techniques by Wang \etal \cite{Wang2019} and Wang \etal \cite{diffWang2020}
    \item Byun \etal \cite{Byun2019} evaluate measures of what they call DNN's ``sentiment'', used to prioritise test inputs that are likely to cause model weaknesses, and they consider if and how SA  achieves the same goal, in addition to other metrics, like confidence score and Bayesian uncertainty. 
\end{itemize}
Despite SA being outperformed as a baseline in such experiments, it remains one of the few reference adequacy criteria for DL systems, since none of these works propose any new adequacy criterion.
In fact, they use SA for purposes other than adequacy (e.g., input prioritisation, selection). 
Even in such scenarios, we expect SA to remain a standard, reference baseline in future research.

\subsection{Papers using SA to validate existing approaches}
A series of papers regard SA as a viable approximation for the usually missing ground truth and assess existing techniques based on their (cor-)relation with SA.
\begin{itemize}
    \item Jahangirova \etal \cite{Jahangirova2020} perform an empirical analysis of various mutation operators and compare their correlations with existing adequacy criteria like SA.
    \item Dong \etal \cite{Dong2019} explore the limited correlation of robustness and coverage in deep neural networks by employing both adversarial robustness metrics and coverage metrics like SA in their work. 
    \item Yan \etal \cite{Yan2020} study the correlations between model quality and coverage criteria, comparing surprise guided retraining with existing adversarial training approaches. 
    \item Chen \etal \cite{Chen2020} evaluate the performance of structural and non-structural coverage criteria on a variety of test sets, shedding more light on the usefulness of these approaches.
\end{itemize}

\subsection{Papers using SA as part of their proposed approach}
These papers use SA as a component of a larger method/technique. 
\begin{itemize}
    \item Li \etal \cite{Li2020} %
    use LSA to calibrate confidence values.%
    \item Kim \etal \cite{Kim2020} use LSA in the domain of question-answering and study the correlation between answer correctness and surprise values.
    \item Ma \etal \cite{Ma2019} use SA values to sort and choose test inputs as candidates for evaluating the performance of a DNN.
    \item Kim \etal \cite{Kim_industry2020}  propose a variant of SA, based on Mahalanobis distance, to reduce labelling costs and apply their technique to an industrial use case in autonomous driving.
\end{itemize} 

Overall, we find that SA is an often used metric, which proved to be useful for  the development of novel approaches, but also as a baseline or as a validation   metric when assessing new DL testing techniques.

\section{Refined Implementation}
\label{sec:implementation}

\begin{listing}
    \begin{lstlisting}[language=Python, numbers=none]
import tensorflow as tf
import LSA

# Prepare dependencies (dataset, trained model)
x_train, y_train, x_test = ... 
model = ...

# SA configuration instance:
config = SurpriseAdequacyConfig(
    saved_path=temp_folder, # Caching folder
    is_classification=True, 
    num_classes=10,
    layer_names=["last_dense"], # Layers used for AT
    ds_name="mnist" # Used for caching
)

# Choose SA subclass (LSA, DSA)
sa = LSA(
    model=model, 
    train_data=train_subset, 
    config=config
)

# Offline SA preparation (here: fit KDE)
# If use_cache is true, cached ATs are used
sa.prep(use_cache=True)

# Calculate the surprises and model predictions
suprises, predictions = sa.calc(
    target_data=x_test, # Test inputs
    use_cache=True, # Allows to re-use previously 
                    # collected ATs (e.g. from DSA)
    ds_type='test' # identifier for caching
)

\end{lstlisting}
\caption{Usage example of our object-oriented interface}
\label{lst:oo-interface}
\end{listing}

Kim \etal\cite{Kim2018} released their implementation of SA under a permissive MIT license, making it easy for the community to use SA in their research.
This was well received: Two years after the release, their github repository is starred 25 times and forked 14 times\cite{Kim2018Code}. 
As a first contribution, we conducted a thorough review and refinement of that source code. 
The primary objectives of our refinements were \textit{performance},
i.e., the online calculation of the SA for given inputs should be fast even if large number of inputs are passed,
and \textit{usability}, i.e., the code should expose a simple, user friendly API.
Towards reaching these objectives, we applied the following techniques:

\noindent
\textbf{Vectorized Processing}
For many common operations, \numpy provides a performance optimized, vectorized implementation.
This implementation tends to scale much better than their equivalent python implementation, which requires to loop over an array.

\noindent
\textbf{Multithreaded Predictions}
We parallelize the calculation of multiple batches of inputs to DSA, thus allowing for a large performance gain on machines with multiple cores.

\noindent
\textbf{One-Shot DNN Forward Passes}
SA requires the calculation of both the DNN prediction as well as the activations of selected hidden layers for any  input.
In the original implementation, this is done using two distinct DNN forward passes.
We replace this with a single forward pass, collecting predictions and hidden layer activations at once, thus saving half of the DNN-prediction cost. 
This is particularly beneficial on systems without GPUs, where DNN evaluations are slow.

\noindent
\textbf{Object Oriented Interface}
For many use-cases, especially in research applications, we expect users to instantiate different types and configurations of SA interchangeably.
To facilitate such usage scenario, our implementation follows an object-oriented architecture, where all variants expose the following common functions:
\begin{itemize}
    \item \textbf{Constructor} The constructor expects a common configuration file containing static information such as: which DNN layers should be used for SA calculation or whether ATs should be cached.
    In addition, it also accepts approach specific parameters (e.g. the batch size to be used when calculating DSA).
    \item \textbf{prepare-method \emph{'prep'}} All offline tasks, i.e., all tasks that have to be performed to allow SA calculation for a novel input, but which do not depend on such novel input, are put into the prepare method.
    Tasks executed in the prepare method include the collection of the ATs of the training set and, for LSA, the fitting of KDE.
    \item \textbf{calculation \emph{'calc'}} The actual calculation of  SA for a given array of DNN inputs.
\end{itemize}

Listing \ref{lst:oo-interface} shows an example usage of our interface.
Note furthermore that we tested our implementation against the original implementation by Kim \etal \cite{Kim2018Code}, verifying that the two implementations remain functionally equivalent.

\section{Refined Approaches}
\label{sec:refined-approaches}

While the improvements of the implementation are a big step towards a faster calculation of DSA, they do not address the underlying algorithmic problem of DSA and LSA: 
In DSA, for a single DNN input, the distance between its AT and \textit{every} AT collected for the training set inputs has to be computed.
Such an approach does not scale well with large training sets, which are particularly common where DNNs are used.
LSA relies on KDE which does not scale with large training sets either\cite{Raykar2010}.

We hypothesize that, as a training set grows, the marginal benefit of an added training input towards improving the quality of SA decreases, 
eventually converging towards zero. 
If our hypothesis is correct, the calculation of SA based on a sampled subset of the training set, $T' \subset T$ with a target sampling ratio $s \in (0,1)$, s.t. $s = \frac{|T'|}{|T|}$, should ideally give comparable results at a runtime cost reduced by factor $s$.
In this paper, we discuss three candidate ways to sample from the training data:
One which aims to preserve the distribution of ATs observed on the full training set, 
one which aims to avoid sampling noise and outliers,
and one which aims to avoid sampling of pairs of nearly-equivalent samples.

\subsection{Distribution-Preserving / Uniform Sampling}
As a first way to sample ATs from the training set, we consider sampling uniformly at random. 
Despite being very simple 
(for a shuffled $T$, it is equivalent to taking the first 
$ s*|T| $ 
entries from the training set) this approach is well motivated:
Uniform-at-random sampling preserves, in expectation, the underlying distribution. 
Thus, with a sample size large enough to represent the training set, we expect SA to perform comparably well.

\subsection{Unsurprising-First Sampling}
Especially for nontrivial classification problems, there will be noise in the training set, which can have a negative impact on OODD capabilities.
This is particularly evident for DSA, which relies on distances to single points in the training set: If a point lies close to noise in the training set, it is unreasonably expected to be unsurprising.
Unsurprising-first sampling aims to sample noise last: 
As low-surprise samples are unlikely to be outliers or noise, sampling only unsurprising training samples aims to prevent the sampling of noise or outliers, and thus facilitate out-of-distribution detection.

Hence, we apply the following sampling procedure, which for every class samples the lowest surprising training data:
Let $C$ denote the set of distinct classes in the considered classification problem and, for every $c\in C$, 
let $T_c$ denote the training samples for which model $M$ predicts class $c$, i.e.,
\[
T_c = \{x | x \in T, M(x) = c\}.
\]
We then choose the subset $T'_c \subseteq T_c$ to contain the $\lfloor s*|T_c| \rfloor$ values of $T_c$ with the lowest surprise 
(we use LSA as metric for surprise due to its general applicability).
Finally, $T' = \bigcup_{c\in C} T'_c$.

\subsection{Neighbor-Free Sampling}
\begin{algorithm}[t]
\SetAlgoLined
\KwResult{A subset of T without $\epsilon$ neighbors}
  \SetKwInOut{Input}{Inputs}
  \SetKwInOut{Output}{Output}
 \Input{Model $M$, Training Set $T$, Neighborhood-Threshold $\epsilon$ }
 \vspace{1mm}
  $T' \gets \emptyset$\;
 \ForEach{distinct class $c$ in $\{M(t)\mid t \in T\}$ }{
  $T_c \gets \{x | x \in T, M(x) = c\}$\;
  \While{$|T_c| > 0$}{
   $i \gets \text{pop any from } T_c$ \;
   
   $T' \gets T' \cup \{ i \} $ \;
   
  $T_c \gets \{x \mid x \in T_c \land \lVert at(x) - at(i) \lVert \geq \epsilon \}$ \;
   }{
  }

 }
 \vspace{1mm}
  \Output{$T'$}
 \caption{$\epsilon$-neighbor free sampling}
 \label{alg:neighbor-sampling}
\end{algorithm}

Neigbor-Free sampling builds on the idea that, given a pair of very similar ATs from the training set, 
we do not expect any large change in DSA values if one of the two inputs from the pair is removed.
As DSA relies only on the distances to the two closest ATs (for same and different class, respectively), 
removing $x_1$ or $x_2$ from the training set does not change these distances if $x_1 = x_2$, while a negligible perturbation is expected if the distance between the ATs of $x_1, x_2$ is smaller than some small~$\epsilon$, i.e.,
$\lVert at(x_1) - at(x_2) \Vert < \epsilon$, still with $M(x_1) = M(x_2)$ and $x_1 \neq x_2$. 
In such a case, we say that $x_1$ and $x_2$ are $\epsilon$-neighbors.
The minimal distance between any sample $x \not\in \{x_1, x_2\}$ and any point in $T$ can be easily shown to increase by at most $\epsilon$ when removing one of $x_1, x_2$. 
Thus, by extension,  the observed DSA values will also change just slightly.
Thus, in $\epsilon$-neighbor-free sampling, we select a subset of the training set $T'_c \subseteq T_c$ such that there are no $\epsilon$-neighbors within $T'$.

We propose a greedy process to perform such sampling, and refer to \autoref{alg:neighbor-sampling} for a definition of this greedy process.
A disadvantage of this algorithm is that it takes $\epsilon$ instead of the target sampling ratio $s$ as input.
Thus, to reach a specific $s$, a suitable $\epsilon$ has to be found.
Furthermore, we acknowledge that the worst-case runtime for this algorithm grows quadratically in the training size.
This problem might be, to some extent, mitigated in practice:
The algorithm runs offline, i.e., it  has to be executed only once, after the training of the DNN is completed and before any SAs for any input have to be computed. 
Moreover, the algorithm is easily parallelizable by handling every class in a different process.
Finally, with large enough $\epsilon$, the expected number of iterations in the inner \textsc{WHILE}-loop is clearly less than the worst-case number of iterations.

While these three sampling strategies are well motivated for DSA, where the calculated surprise metric is dependent on the closest points for the same and different class prediction,
for LSA, which bases its calculation on the complete distribution of ATs, we consider only distribution-preserving sampling as a valid sampling strategy. 
Outlier removal (as performed by unsurprising-first sampling) is not required for LSA, as outliers have  low density by definition.
$\epsilon$-neighbor-free sampling would reduce the density in the most dense regions, thus defeating the point of using a density estimator in LSA.
\section{Empirical Evaluation}
\label{sec:assessment}

\def \unctypesubfigurewidth {0.2\linewidth}
\def \unctypeimgwidth {.95\linewidth}

\begin{figure}[t]
\centering
\begin{subfigure}{\unctypesubfigurewidth}
  \centering
  \includegraphics[width=\unctypeimgwidth]{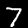}
  \caption{}
  \label{fig:ds_nominal}
\end{subfigure}
\begin{subfigure}{\unctypesubfigurewidth}
  \centering
  \includegraphics[width=\unctypeimgwidth]{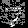}
  \caption{}
  \label{fig:ds_adv}
\end{subfigure}
\begin{subfigure}{\unctypesubfigurewidth}
  \centering
  \includegraphics[width=\unctypeimgwidth]{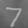}
  \caption{}
  \label{fig:ds_corr}
\end{subfigure}%

\caption{
Examples of the datasets used for our OODD experiments:
(\subref{fig:ds_nominal}) shows a nominal image
(\subref{fig:ds_adv}) shows its adversarial (model specific) modification
(\subref{fig:ds_corr}) shows its modification using a realistic corruption.
}
\label{fig:ds_types}
\end{figure}

We conducted an empirical study to answer the  following research questions:

\noindent
\textbf{RQ\textsubscript{0} (Reproduction):}
\textit{Is SA capable of performing OODD, as reported when first introduced?}

Kim \etal\cite{Kim2018} report, amongst other results, OODD capability of SA for MNIST.
Our RQ\textsubscript{0} is a partial reproduction of their results, however using a different model, different outlier datasets, and our refined implementation of plain SA.

\noindent
\textbf{RQ\textsubscript{1} (Evaluation Time):}
\textit{To what extent do our performance-optimized implementations and refined approaches increase SA calculation speed?}

We measure the time improvement on calculating both LSA and DSA using our performance-optimized approaches over the original implementation of SA \cite{Kim2018Code}.
We furthermore measure the speed improvement when using an increasingly smaller sample of ATs from the training set used in SA,
to verify that sampling is indeed an appropriate strategy to improve SA calculation runtime.

\noindent
\textbf{RQ\textsubscript{2} (Detection Capability of Refined Approaches):}
\textit{How effective are our refined variants of SA to detect adversarial and corrupted inputs?}

We have to expect an influence of the sampling of the training set in our refined SA approaches on the associated OODD capability.
We thus measure the OODD performance under various threshold sizes on two different outlier datasets.
This allows us to assess if and when indeed our refined sampling approaches are still valid  for OODD.

\noindent
\textbf{RQ\textsubscript{3} (Retraining Instability):}
\textit{How sensitive are the different SA approaches to random influences due to the training process?}

Training the same model multiple times  typically leads to different models -- even when the resulting model performance (e.g. accuracy) is comparable.
Amongst other reasons, this is due to random weight initialization and to the presence of any layers with a random behavior (such as Dropout layers).
Slightly different weights will also lead to slightly different ATs for the same input, which could impact the values of SA.
We conducted our experiments on 100 models with the same architecture, but  trained independently, allowing us to not just elicit the expected (i.e., average) OODD performance of SA, but also to measure its \emph{training instability} i.e., the sensitivity of SA to the non-deterministic training process.

\subsection{Experimental Setup}
We use MNIST \cite{LeCun1998}, a dataset consisting of 70'000 (60'000 training and 10'000 test) samples of grayscale handwritten digits.
It is the most widely used dataset in testing of deep learning based software systems\cite{Riccio2020}.
We use a DNN architecture consisting of two convolutional (and max-pooling) layers, followed by a dropout and two dense layers.

To measure the OODD capabilites, we generate two Out-of-Distribution datasets: 
First, we use an adversarial dataset, where the MNIST test set is modified using the \emph{fast gradient sign method (FGSM)} provided in the \emph{foolbox} library\cite{
Rauber2017}.
Such data is model-specifically modified to lead to mispredictions.
Adversarial data, including some generated using FGSM, were used to assess SA when it was first released.
Second, we use a \emph{corrupted} dataset, based on corruptions provided in Mu \etal \cite{Mu2019}.
The applied changes aim to modify the test data using realistic corruptions, such as changing contrast or inserting additional elements on the picture.
An example of a nominal input (from the MNIST test set) and its modified out of distribution versions (adversarial and corrupted) are shown in \autoref{fig:ds_types}.

We assess the OODD capabilities of SA by calculating the surprise of both the full nominal dataset as well as the outlier dataset.
Individually, we then measure how well SA allows to discriminate between the nominal data and the outliers, using the AUC-ROC score. 
Note that for this analysis, in line with the original study on SA \cite{Kim2018}, whether a test input is correctly classified has no influence on the collected AUC-ROC score, as the task to perform is not to distinguish between correctly classified and misclassified inputs, but to distinguish between nominal and modified data sources.
Using an existing testing tool\cite{Weiss2021failsafe}, we ran our experiments on 100 independently trained models.
This allowed us to mitigate the influence of randomness of the training phase in RQ\textsubscript{0} and RQ\textsubscript{2}, as we report average AUC-ROCs,
and to measure the influence of non-determinism on the SA OODD performance in RQ\textsubscript{3}. 

\subsection{Results}
In this section, we present the key findings of our empirical evaluation by research question.

\subsubsection{RQ\textsubscript{0} (Reproduction)} 
\begin{table}[t]
\begin{tabular}{@{}cccccccc@{}}
\toprule
                             &     & mean  & std   & median & min   & max   & range \\ \midrule
\multirow{2}{*}{Adversarial} & DSA & 
0.971 & 0.008 & 0.974 & 0.046 & 0.988 & 0.023
\\\vspace{1mm}
& LSA & 0.924 & 0.014 & 0.925 & 0.882 & 0.963 & 0.081 \\
\multirow{2}{*}{Corrupted}   & DSA & 0.835 & 0.003 & 0.836 & 0.828 & 0.842 & 0.014 \\
                             & LSA &0.779 & 0.008 & 0.779 & 0.758 & 0.795 & 0.037 \\ \midrule
\end{tabular}
\caption{AUC-ROC computed on 100 re-trained models}
\label{tab:auc_roc_stab}
\end{table}
While, originally, SA was only evaluated on adversarial outliers, we also evaluate it on corruption-based outliers, i.e., outliers created by realistically modifying the test dataset.
The average AUC-ROCs for OODD, measured using 100 different models, were
0.97 on adversarial outliers and 0.84 on corrupted outliers for DSA;
0.92 and 0.78 for LSA.
As shown in \autoref{tab:auc_roc_stab}, the lowest of all four mean AUC-ROCs over both outlier types and SA variants was 0.779.
Since the expected performance of a random OODD on our balanced dataset is 0.5, 
both LSA and DSA are capable of performing accurate OODD, thus reproducing the corresponding results by Kim \etal\cite{Kim2018} (their RQ\textsubscript{1}).

\begin{tcolorbox}
\textbf{Summary (RQ\textsubscript{0})}: 
Our experiments are consistent with the findings by Kim \etal
and show that both LSA and DSA are effective at OODD,
i.e., they capture the degree of surprise of the inputs.
\end{tcolorbox}

\subsubsection{RQ\textsubscript{1} (Evaluation Time)} 
We evaluate performance improvements of our SA implementation by measuring the time required to calculate SA on the complete MNIST test set (10'000 inputs).
This evaluation deliberately ignores offline preparation of SA (such as fitting KDE), which only have to be computed once per model. %
We  considered the  runtime on two different machines: 
A MacBook-Pro Notebook (2019, 2.6Ghz i7 CPU) and a Linux Desktop-PC (AMD Threadripper 1920X CPU and a RTX 1070-TI GPU).
For DSA, we report the performance improvements both using single- and multithreaded DSA batch predictions,
to allow a fair comparison with the original implementation, which is single-threaded.
Performance was measured on only one model per approach (i.e., the experiment was only run once), and the machine was not used for any other tasks during that time.
As we ignore offline preparation of SA time, the measured runtimes are independent on the sampling method.%

Our results are reported in \autoref{tab:runtimes}.
The performance increases are most apparent for DSA. 
Vectorized processing reduces the cost to $\sim$25s on both machines when executed without multithreaded batch processing, where the time on the original implementation was 4:44min (MacBook) and 2:12 (Desktop-PC). 
Processing multiple batches in parallel further reduced time to 13s or 4s, respectively.
The relative change is largest for the Desktop-PC: Our multithreaded implementation is 33 times faster than the original DSA implementation.
The improvement is less dramatic for LSA:
On the Desktop PC, probably due to the different OS, KDE is slower anyway and it also appears not to be able to perform faster when given batches of inputs. 
On the MacBook however, performance improved from 10s to 4s.
It is also notable that, while DSA is clearly slower than LSA in the original implementation, 
for MNIST-sized problems, our multithreading version of DSA makes it the faster of the two approaches.

\begin{figure}[t!]
    \centering
    \includegraphics[width=\linewidth]{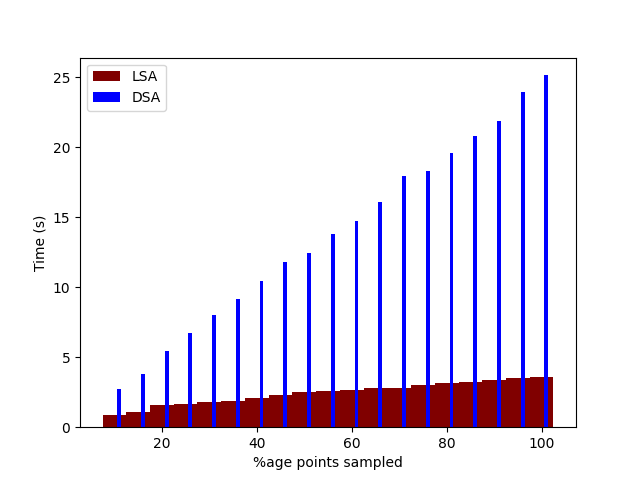}
    \caption{Prediction times on the MacBook with different sampling sizes. DSA was run without batch multithreading. }
    \label{fig:sampling-times}
\end{figure}

We furthermore verified our expectations on the performance impact of  subsampled training sets:
The results, measured on the MacBook Pro are shown in \autoref{fig:sampling-times}. 
As expected, for both LSA and DSA, computation costs are decreasing with smaller training set samples.
For DSA we furthermore observe the runtime to be proportional to the training set sample size,
but also for LSA, the improvement is clear: 
while calculating LSA using 100\% of the training data takes 4.24s, using a subsample of 10\% of the data reduces training time to 0.84s (80.2\% improvement).

Some of these absolute time improvements may appear too small to be relevant in practice, in particular for LSA.
However, it is worth noting that SA may be used as part of a repetitive test procedure, 
e.g. in search based test set generation, where the test set quality is assessed using SA.
Here, even small absolute improvements on a single SA prediction can build up to large time savings in the overall test procedure.

\begin{tcolorbox}
\textbf{Summary (RQ\textsubscript{1})}: 
Our fast implementation of SA, which is functionally equivalent to the original implementation,
reduced SA calculation cost heavily: up to 97\% for DSA and 60\% for LSA.
Sampling-based SA can further increase SA calculation speed.
\end{tcolorbox}

\begin{table}[t]
\centering
\begin{tabular}{@{}ccccccc@{}}
\toprule
           & \multicolumn{2}{c}{LSA} && \multicolumn{3}{c}{DSA}                                                                                                                        \\ \cmidrule{2-3}\cmidrule{5-7}
           & orig.      & impr.   &   & orig. & \begin{tabular}[c]{@{}c@{}}impr. (ST) \end{tabular} & \begin{tabular}[c]{@{}c@{}}impr. (MT) \end{tabular} \\
           \midrule
MacBook Pro   & 0:10 &  0:04         &            &  4:44     &   0:24                                                                 &  0:13                                                                 \\
Desktop-PC & 0:12 & 0:12           &          & 2:12 &    0:25  &         0:04                                                                                                                             \\ \bottomrule

\end{tabular}
\caption{Runtime comparison in mm:ss of our improved (impr.) and the original (orig.) SA implementation. (\emph{ST/MT} = Single/Multi-threaded batch processing)}
\label{tab:runtimes}
\end{table}

\subsubsection{RQ\textsubscript{2} (Detection  Capability  of  Refined  Approaches)} 

\def \subfigurewidth {0.9\linewidth}
\def \imgwidth {.5\linewidth}

\begin{figure*}[t!]
\begin{minipage}[b]{\imgwidth}
\centering
\includegraphics[width=\subfigurewidth]{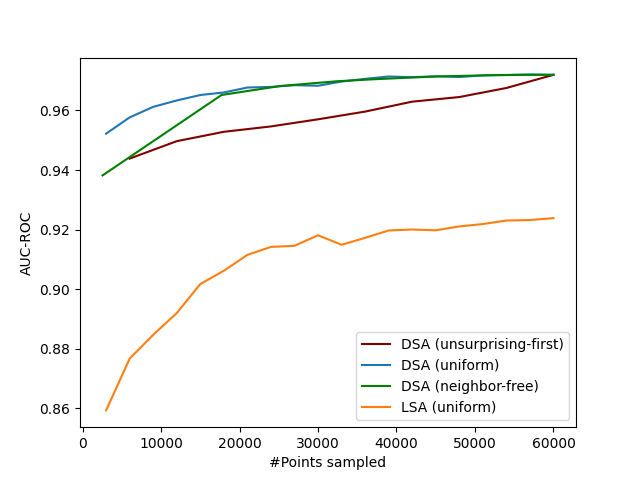}
\subcaption[]{Adversarial Outliers}
\label{fig:oodd-sampled-adversarial}
\end{minipage}
\hspace{0.5cm}
\begin{minipage}[b]{\imgwidth}
\centering
\includegraphics[width=\subfigurewidth]{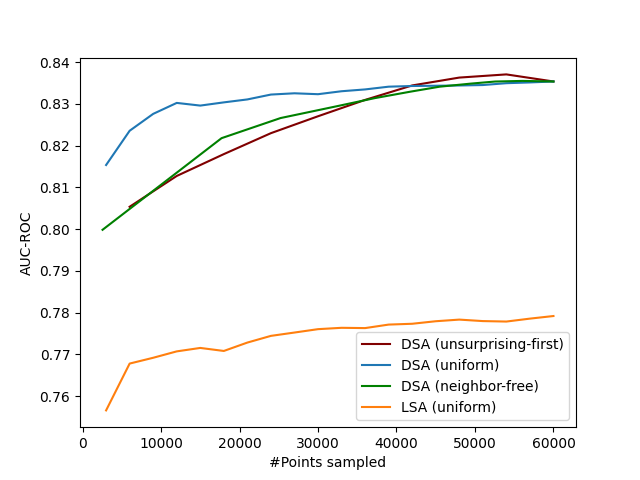}
\subcaption[]{Corrupted Outliers}
\label{fig:oodd-sampled-corrupted}
\end{minipage}
\caption{Out-of-distribution detection performance with varying sample sizes and sampling strategies}
\label{fig:ood-sampled}
\end{figure*}

To plot the OODD capability at increasing sample size we applied the following procedure:
We chose uniformly distributed steps of $s$ for distribution-preserving and unsurprising-first sampling.
For $\epsilon$-neighbor-free sampling, we estimated the steps of $\epsilon$ leading to an approximately uniform distribution of $s$ based on sampled distances from the training set.
At every step we then averaged, over all 100 re-trained models, the  AUC-ROCs.
The resulting plots are shown in \autoref{fig:oodd-sampled-adversarial} (adversarial outliers) and \autoref{fig:oodd-sampled-corrupted} (corrupted outliers).

\paragraph{DSA} Using a sample ratio of $\frac{1}{3}$, the capability to detect adversarial samples does not notably decrease (less than 1\%) when using either uniform or $\epsilon$-neighbor-free sampling. For unsurprising-first sampling, the decrease is slightly larger, but still less than 2\%. 
While all sampling strategies perform well in detecting corrupted inputs, uniform sampling performs slightly better then the other two strategies -- which is notable since this is also the simplest and fastest sampling strategy.

\paragraph{LSA} The OODD performance of LSA appears slightly more impacted by  sampling than DSA, in particular on the adversarial outliers.
Nonetheless, we still observe good AUC-ROCs at a sampling rate of $\frac{1}{2}$, thus motivating the use of sampling when the training set becomes too big also with LSA.

\begin{tcolorbox}
\textbf{Summary (RQ\textsubscript{2})}: 
In our case study, even when considering only 33\% of the training data for computing DSA, or 50\% with LSA, 
OODD capability decreases only minimally.
\end{tcolorbox}

\subsubsection{RQ\textsubscript{3} (Training Instability)} 
We used the 100 independently but equivalently, trained models to measure the impact of non-determinism on SA's OODD capability and denote such sensitivity as \emph{Training Instability}.
We measure the training instability using the \emph{standard deviation (std)} of the  AUC-ROC values, as well as their \emph{range}, i.e., the difference between the highest and the lowest observed AUC-ROC. 
Our results are shown in table \autoref{tab:auc_roc_stab}.
We observe that training instability, in particular for LSA, is high enough to be possibly relevant in practical use-cases.

\begin{figure}[t]
    \centering
    \includegraphics[width=0.95\linewidth]{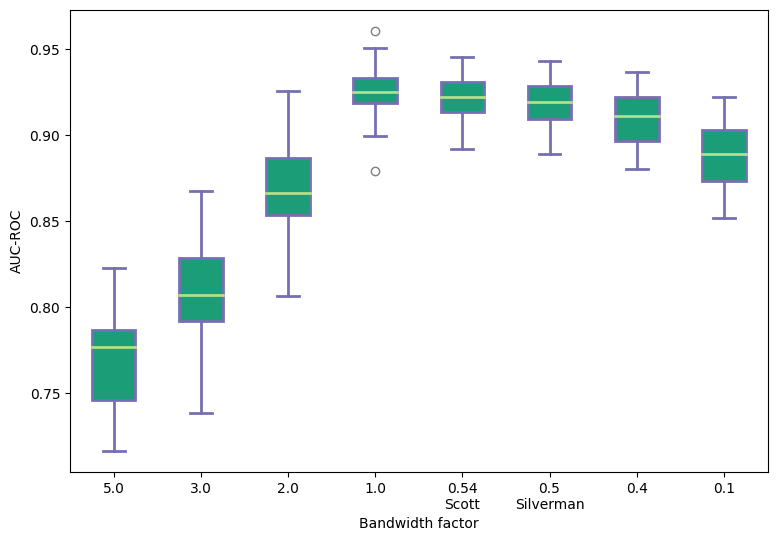}
    \caption{Distribution of OODD AUC-ROC using the adversarial outliers for selected KDE bandwidths}
    \label{fig:lsa-bandwidth}
\end{figure}

KDE, and thus LSA depend on one important hyperparameter: the bandwidth used when fitting KDE. 
To see whether we can use this hyperparameter to reduce training instability, we ran LSA to perform OODD using the adversarial datasets under a selection of different bandwidths.
The resulting AUC-ROC distributions for various bandwidths are shown in \autoref{fig:lsa-bandwidth}.
They confirm that indeed, bandwidth selection has a high impact on the distribution of AUC-ROCs, as apparent from the variability of the inter-quartile range across bandwidths.
\emph{Heuristics} such as \emph{Scott's Rule} (default in LSA) or \emph{Silverman's Rule} provide good results, ensuring low instability in our case study.

\begin{tcolorbox}
\textbf{Summary (RQ\textsubscript{3})}: 
In our case study, LSA shows higher training instability than DSA. 
To get reliable results, LSA should be computed on multiple, independently trained DNN models, thus allowing to observe LSA value distributions and to compute their mean value.
Moreover, the choice of bandwidth in KDE directly influences OODD performance as well as training instability. %
\end{tcolorbox}

\subsection{Threats to Validity}
The primary threat to validity of our study is the low number of test subjects:
While we considered two structurally different types of outliers, we use only one underlying dataset: MNIST.
Furthermore, we acknowledge that MNIST is a comparably small and clean dataset, which may further hinder generalizability of our results for real world, large and noisy datasets.
Note however that this threat does not apply to all our results:
First, the fact that sampling-based DSA is much faster is due to analytical properties of sampling that generalize to other datasets.
Furthermore, to raise the issue that LSA \emph{can be} sensitive to random influences during training, one example showing such sensitivity is sufficient. 

\section{Future Work}
\label{sec:future-work}

Reproduction of our experiments on different datasets, different models, and different layers used for AT collections are required to generalize our findings on the OODD capability of sampling based DSA and to support the choice of sampling method.
In particular, future studies should consider datasets which are currently harder to be used for OODD, such as noisy and heavily unbalanced datasets.
Furthermore, our SA refinements should be compared to existing refinements of SA, in particular Mahalanobis based SA\cite{Kim_industry2020}.

More experiments on other datasets are required to further enlighten the training instability of SA and to confirm or reject the hypothesis that LSA is typically more sensitive to random influences during training than DSA.
\section{Conclusion}
\label{sec:conclusion}

A major weakness of SA is that SA calculation does not scale well with large training datasets.
We mitigate this weakness by providing both a faster implementation of plain SA and by suggesting a performance improvement of SA through training set sampling.
In our benchmark experiments, our plain DSA implementation was up to 33 times faster than the implementation of DSA released upon its first publication.
On MNIST, our sampling strategies further halved the runtime, at  a very small cost in terms of reduced OODD AUC-ROC (1\%).
Our experiments also raise the concern of training instability in SA, i.e., a high sensitivity of its performance to random influences during training.

\section{Data Availability}
\label{sec:data}
We release the reproduction package of our paper under a permissive MIT license\footnote{
\href{https://github.com/testingautomated-usi/surprise-adequacy}{github.com/testingautomated-usi/surprise-adequacy}
}.

\typeout{}
\bibliographystyle{IEEEtran}
\bibliography{main}

\end{document}